# Comparative evaluation of photogrammetric reconstruction methods and 3D Gaussian Splatting for road surface roughness analysis

Marouane Elmegdar[a,b], Teng Xiao*[b]
[a] School of International Education, Hubei University of Technology, Wuhan, China 430068;
[b] School of Computer Science, Hubei University of Technology, Wuhan, China 430068;
elmegdarmarouane@gmail.com, *xiao@hbut.edu.cn

## ABSTRACT

Image-based 3D reconstruction offers a low-cost alternative to traditional sensor-based techniques for road surface assessment. This study compares four reconstruction pipelines—COLMAP, Meshroom, Metashape, and 3D Gaussian Splatting (3DGS)—to evaluate their ability to estimate road surface roughness from smartphone imagery. All point clouds were processed in CloudCompare using a consistent workflow involving orientation alignment, segmentation, normal estimation, and roughness computation at neighborhood radiuses of 0.2, 0.4, and 0.6 model units. The results show that COLMAP provides the highest sensitivity to micro-texture, while Meshroom yields balanced reconstructions with moderate roughness variation. Metashape produces the smoothest geometry due to its internal filtering, and 3DGS captures visible irregularities but exhibits higher noise and lower density. The comparison demonstrates that open-source pipelines are viable for relative roughness evaluation, offering a practical approach for low-cost pavement monitoring.

## 1. INTRODUCTION

Road surface roughness is a key indicator of pavement health and affects driving comfort, vehicle maintenance, and road safety [1]. Traditional inspection methods such as profilometers and laser-based inertial systems provide high-accuracy measurements, but they require specialized hardware, trained operators, and costly deployment [1,2]. Their use is therefore limited to large-scale or periodic inspections.

Image-based 3D reconstruction offers a low-cost and scalable alternative by enabling geometric surface evaluation from consumer-grade cameras [3]. Photogrammetric Structure-from-Motion (SfM) and Multi-View Stereo (MVS) techniques have been widely applied for road monitoring, showing strong potential for detecting irregularities and surface degradation [4,5]. Open-source tools such as COLMAP and Meshroom are attractive due to accessibility and flexibility, though their performance can be influenced by illumination, surface texture, and processing parameters [3]. Commercial solutions like Agisoft Metashape provide enhanced robustness and cleaner surfaces but remain proprietary and less aligned with low-budget deployment in transportation projects.

Recently, radiance-field-based reconstruction approaches such as 3D Gaussian Splatting (3DGS) have gained rapid attention due to their ability to generate visually coherent and dense reconstructions with efficient processing [6]. However, prior studies have primarily evaluated these methods on rendering performance and scene completeness, with limited focus on metric reliability for road surface roughness. Furthermore, direct comparisons between open-source SfM/MVS, commercial photogrammetry, and radiance-field representations under identical conditions remain scarce. Key challenges — scale normalization, geometric noise, varying point densities, and parameter sensitivity — are still not fully addressed in the literature [7].

To fill these gaps, this work provides a unified quantitative comparison of four reconstruction pipelines — COLMAP, Meshroom, Metashape, and 3DGS — using a single smartphone dataset processed with standardized roughness computation in CloudCompare. The objective is to evaluate how reconstruction strategies influence roughness estimation across different neighborhood scales and to identify practical low-cost solutions for pavement monitoring.

## 2. METHODOLOGY

### 2.1 Overview

This study was designed to evaluate how different 3D reconstruction approaches can estimate road surface roughness from image-based models. Four reconstruction tools were compared: COLMAP, Meshroom, Agisoft Metashape, and 3D Gaussian Splatting (3DGS). All reconstructions were analyzed within CloudCompare through the computation of roughness and other relevant metrics.

The workflow followed a standardized structure similar to recent studies on multi-method 3D comparison [7], and included four main phases: (1) image acquisition, (2) 3D reconstruction, (3) point cloud alignment and preprocessing in CloudCompare, and (4) roughness computation and analysis. This framework ensured reproducibility, consistency, and fair evaluation of all reconstruction pipelines under identical conditions, see Figure 1.

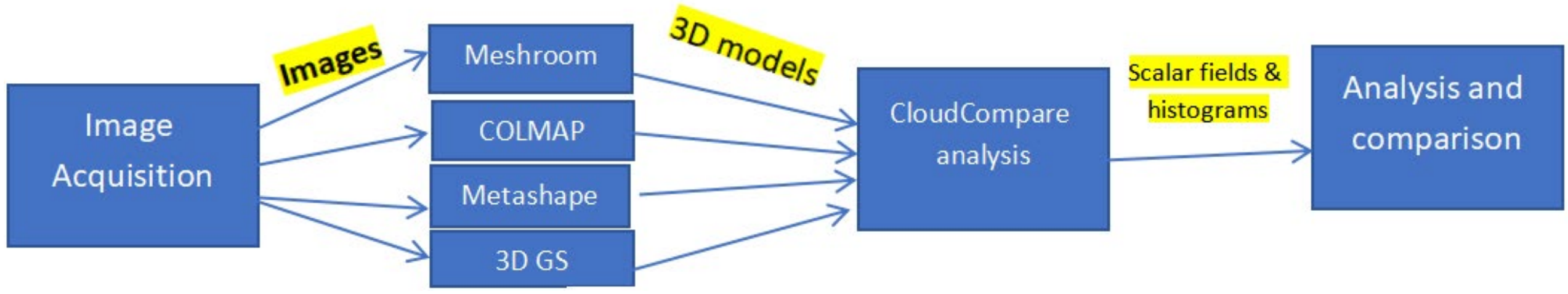

Figure 1. Experiment workflow.

### 2.2 Image acquisition

A road segment was captured using a handheld smartphone camera under stable daylight conditions to minimize shadows and illumination differences, see Figure 2. The images were taken at a consistent height with approximately 70% overlap and camera angles ranging from 30° to 40° to ensure sufficient parallax for feature matching and complete coverage of the surface.

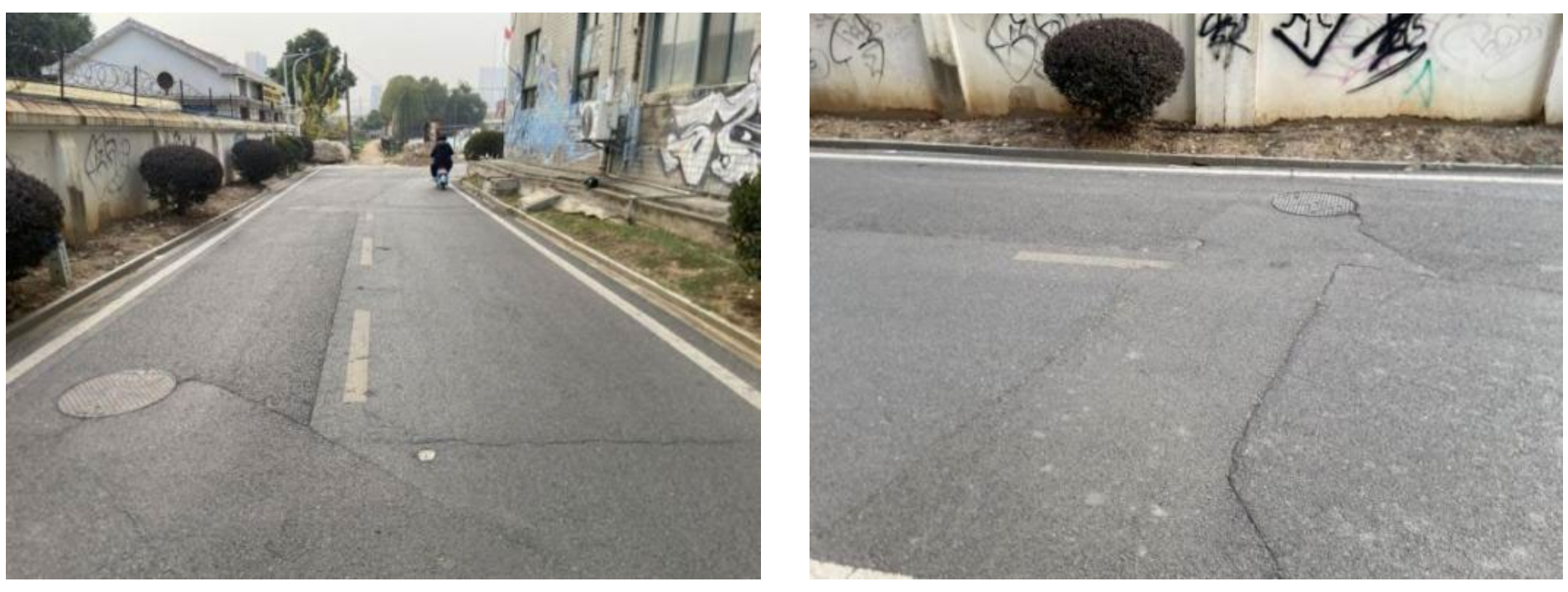

Figure 2. Images of the road segment.

### 2.3 Reconstruction pipelines

The selected reconstruction methods represent three methodological classes: open-source SFM/MVS (Colmap/Meshroom), a commercial photogrammetric solution (Agisoft Metashape) and a radiance field-based representation approach (3D Gaussian splatting). All using identical input images and executed using default parameters to avoid subjective tuning, and their resulting point clouds were exported for analysis.

**COLMAP**[8] performs feature-based Structure-from-Motion and dense stereo reconstruction, followed by depth-map fusion. Its workflow preserves micro-geometry with minimal smoothing, making it suitable for examining high-detail surface variations. Within this study, COLMAP represents a high-detail / low-smoothing baseline.

**Meshroom**[9] uses the AliceVision pipeline to perform depth map estimation, filtering, and meshing [10]. Its default configuration produces reconstructions that balance detail preservation and noise suppression. This positions Meshroom as an intermediate photogrammetric reference between the highly detailed COLMAP output and the smoother Metashape surface.

**Agisoft Metashape**[11] automates image alignment, depth-map generation, and dense cloud interpolation using optimized internal filtering. Its reconstructions tend to be visually smooth and geometrically consistent, making it representative of a commercial, high-stability / high-smoothing pipeline frequently used in applied surface modeling.

**3D Gaussian Splatting**[12] models the scene as a set of Gaussian primitives optimized by radiance and density parameters. After training, the optimized splats were converted to a point cloud for roughness computation. Because 3DGS prioritizes visual coherence instead of explicit geometric accuracy, it often produces non-uniform densities and smoother or scattered surface patches. It serves here as an example of a radiance-field-based reconstruction used for evaluating whether such representations can support reliable geometric roughness estimation.

### 2.4 Hardware requirements and processing time

The experiments were conducted on a computer equipped with an AMD ryzen 7 7800X-3D 8-core, 32GB of memory and an NVIDIA RTX 4070 SUPER GPU with 12GB VRAM. Processing a dataset of 161 images with a resolution of: 4032 x 3024 pixels and each pipeline behaved as follows, see table 1.

Table 1. Pipelines effeciency.

| Tool | Memory Usage | GPU usage | Processing time |
|---|---|---|---|
| **Colmap** | 24GB | 20% | 62 min |
| **Meshroom** | 9GB | 98% | 85 min |
| **Metashape** | 15GB | 90% | 36 min |
| **3DGS** | 16GB | 100% (high GPU required) | 46 min |

### 2.5 CloudCompare processing

All point clouds were aligned to a common orientation, manually segmented to isolate the road surface, and normalized to consistent model units. Local roughness was computed from the mean absolute distance of points to a best-fit plane within neighborhood radius parameters of 0.2, 0.4, and 0.6 model units. The roughness at point $p$ is:

$$R_p=\frac{1}{n}\sum_{i=1}^{n}\left|d_i-\bar{d}\right| \quad (1)$$

Where $d_i$ is the orthogonal point-to-plane distance, $\bar{d}$ is the mean of these distances and n is the number of neighbors.

Surface roughness was evaluated at multi-scale neighborhood radius of 0.2, 0.4, and 0.6 model units.

In CloudCompare, the neighborhood radius defines the spatial extent of points used to fit the local best-fit plane. Smaller radius captures micro-texture and noise, while larger radius reveals smoothed macro-level surface trends. Evaluating multiple radius enables analysis of how each reconstruction method preserves roughness at different spatial scales.

## 3. RESULTS

This section evaluates the roughness behavior of the four reconstruction pipelines at three neighborhood radiuses (0.2, 0.4, and 0.6 model units). For each radius, the scalar-field visualizations illustrate the spatial distribution of surface deviations, while the numerical means summarize the overall roughness magnitude. The results aim to assess how each pipeline preserves geometric details at increasing neighborhood scales. A detailed analysis is presented followed by corresponding figures.

Using a neighborhood radius of 0.2 model units, see figure 3, COLMAP produced the highest micro-surface variation, capturing fine irregularities and noise. 3DGS also showed high variation but with a more scattered, non-uniform pattern. Meshroom preserved noticeable details while suppressing excess noise, whereas Metashape appeared the smoothest due to stronger internal filtering. The mean roughness values reflected these trends as well: COLMAP had a mean of 0.0225, 3DGS with 0.0268, Meshroom with 0.0064 and Metashape with 0.0050.

At a radius of 0.4 model units, see figure 4, the roughness patterns became more stable as micro-variations were averaged out. COLMAP continued to show the strongest surface variation, 3DGS became slightly more homogeneous, Meshroom maintained a balanced reconstruction, and Metashape remained the smoothest. The mean values increased proportionally with the radius: COLMAP with 0.0410, 3DGS with 0.0366, Meshroom with 0.0114 and Metashape with 0. 0066.Meaning that increasing the radius led to more smoothed roughness responses across all reconstructions.

At 0.6 model units, the scalar fields reflected broader, macro-level surface undulations, see figure 5. COLMAP still retained high-variation zones, 3DGS became more stabilized compared to smaller radius, Meshroom continued producing consistent texture, and Metashape showed minimal variation. The means kept the same ranking: COLMAP with 0.0602, 3DGS with 0.0405, Meshroom with 0.0158 and Metashape with 0.0079, indicating that the neighborhood averaging at 0.6 filtered out all the small-scale fluctuations.

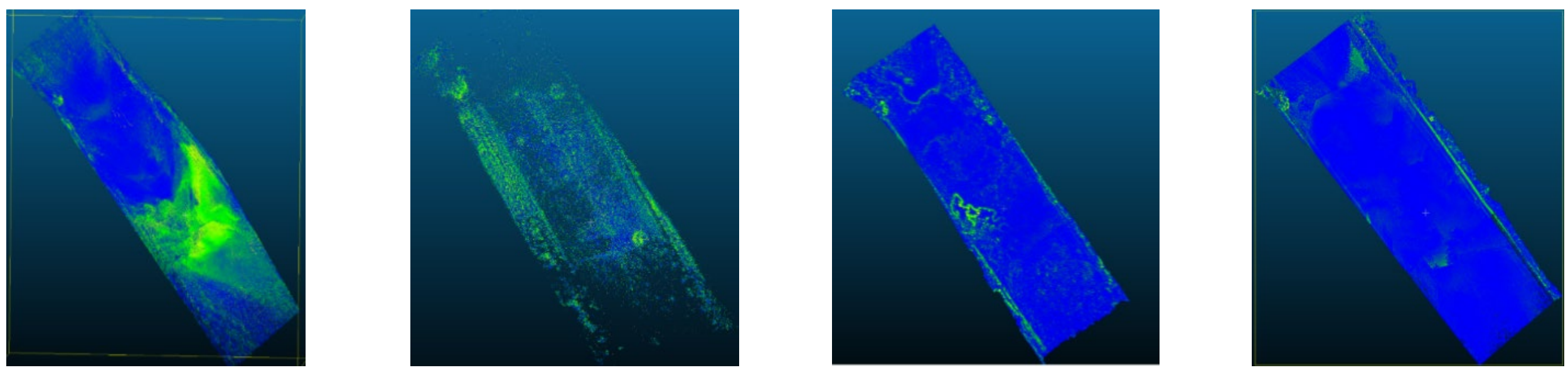

Figure 3. Scalar fields on 0.2 model units radius of: (a) Colmap, (b) 3DGS, (c) Meshroom and (d) Metashape.

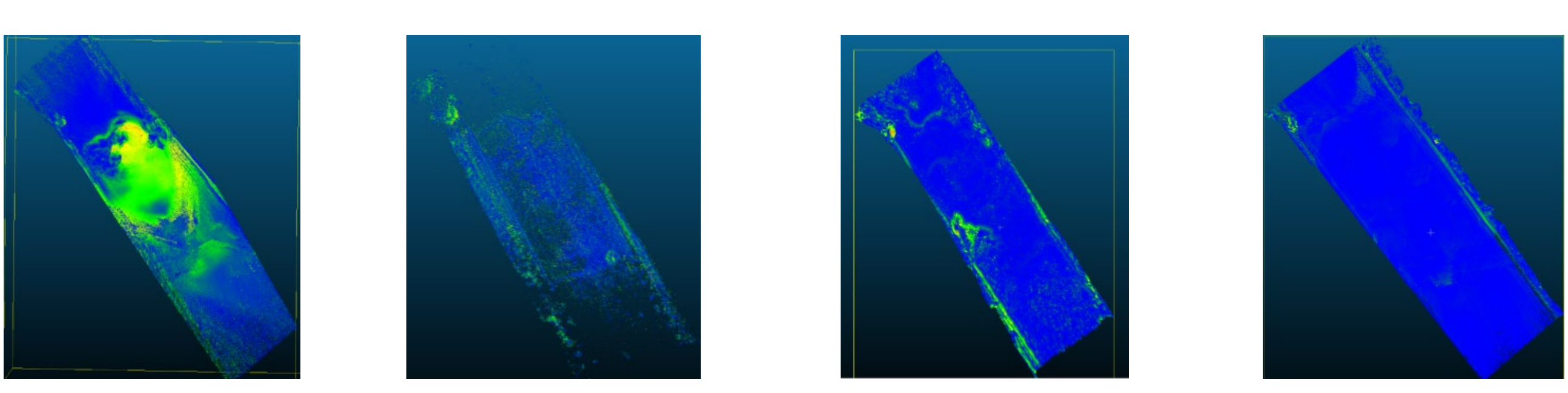

Figure 4. Scalar fields on 0.4 model units radius of: (a) Colmap, (b) 3DGS, (c) Meshroom and (d) Metashape.

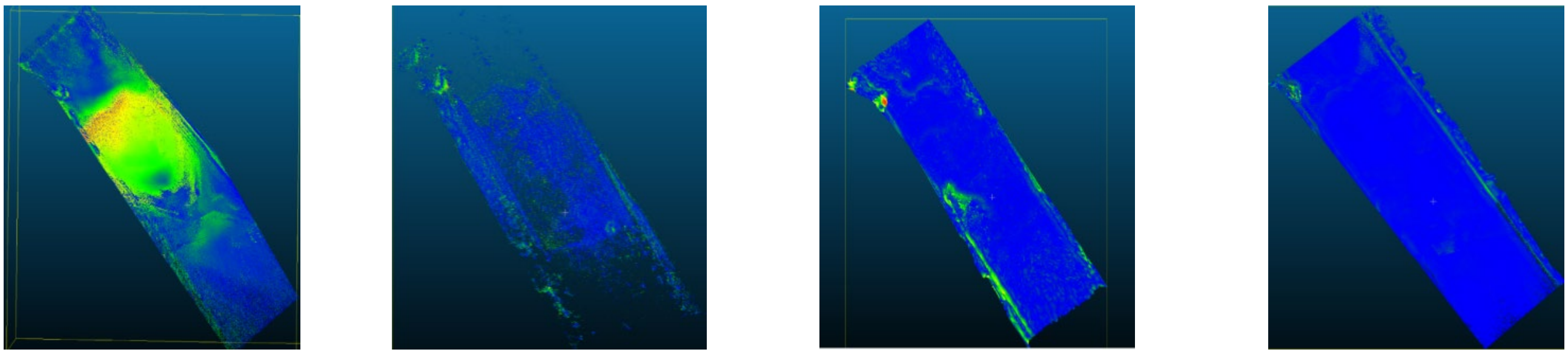

Figure 5. Scalar fields on 0.6 model units radius of: (a) Colmap, (b) 3DGS, (c) Meshroom and (d) Metashape.

## 4. DISCUSSION AND CONCLUSION

This comparative evaluation of four 3D reconstruction approaches highlights their differing behaviors for road-surface roughness assessment, as shown in table 2.

At a smaller radius, both COLMAP and 3DGS captured pronounced and clear micro-surface irregularities, revealing cracks, edges, and fine-scale pavement texture; however, they were also exposed to amplified noise. 3DGS, in particular, exhibited scattered roughness values, consistent with its probabilistic Gaussian representation and spatial sparsity. Meshroom produced a more balanced reconstruction, preserving fine surface variation while effectively suppressing excessive high-frequency noise. In contrast, Metashape generated the smoothest surfaces due to its strong internal filtering and dense point interpolation. These trends were confirmed by the roughness mean values as well.

As the analysis radius increased (0.4 and 0.6 model units), surface roughness gradually stabilized across all models due to spatial averaging. COLMAP and Meshroom retained visible texture gradients and realistic undulation patterns. Metashape's smoothing behavior became dominant at larger scales, resulting in minimal surface variance, whereas 3DGS continued to exhibit slightly dispersed responses in defect zones such as cracks or depressions. These observations confirm that the roughness metric ($R_p$) effectively captures both micro- and macro-geometric characteristics across reconstruction pipelines, providing a reliable quantitative basis for comparative evaluation.

The observed differences among pipelines can be attributed to key algorithmic choices: (1) feature matching density and triangulation quality (affecting local geometric fidelity), (2) depth-map filtering and interpolation that reduces micro-variation in commercial tools, and (3) representation model (triangulated points/meshes vs. probabilistic splats). COLMAP's conservative fusion preserves high-frequency geometry but can amplify measurement noise; Metashape's depth interpolation and internal smoothing suppress small-scale roughness, yielding lower $R_p$ means; Meshroom behaves intermediate due to moderate filtering; and 3DGS's density and radiance-driven optimization can produce coherent renderings whose point/splat distribution increases variance in computed geometric descriptors. Therefore, differences in $R_p$ reflect both true geometry capture and algorithmic smoothing/noise amplification.

Overall, the experiments validate the feasibility of using image-based 3D reconstruction pipelines for road surface roughness analysis, supporting findings from prior photogrammetric studies. Compared to traditional sensor-based approaches, the evaluated methods offer a non-contact, low-cost alternative capable of producing geometrically consistent roughness estimates. Among the photogrammetric tools, COLMAP delivered the highest geometric sensitivity but suffered from local noise, Meshroom achieved the best trade-off between detail and stability, and Metashape provided visually smooth yet metrically conservative results. 3DGS demonstrated potential for geometric reconstruction, but its current implementation remains limited by irregular density and incomplete surface consistency, constraining its use for metric evaluation.

Table 2. Summary of comparative roughness analysis across reconstruction pipelines.

| Tool | Mean (0.2–0.6 model units) | Scalar Field (SF) Appearance |
|---|---|---|
| **Colmap** | 0.0225–0.0602 | Highly detailed with micro-variation; retains fine texture but exhibits noticeable noise. |
| **3DGS** | 0.0267–0.0405 | Scattered, less dense geometry; captures large-scale roughness but with irregular splat distribution. |
| **Meshroom** | 0.0064–0.0158 | Balanced reconstruction; preserves moderate detail while suppressing noise. |
| **Metashape** | 0.0050–0.0078 | Very smooth, homogeneous surfaces; fine irregularities filtered out. |

Scientifically, these outcomes indicate that algorithmic factors such as feature-matching density, depth filtering, and representation models directly influence the derived roughness ($R_p$) values. Photogrammetric algorithms rely on triangulated geometry, which captures detailed but discrete surface information [13], whereas radiance field–based methods such as 3DGS depend on probabilistic splat optimization, which prioritizes visual coherence over precise spatial accuracy. This fundamental distinction explains the observed differences in geometric roughness among the methods.

Future work should focus on improving scale normalization and automatic alignment among reconstructed models to ensure consistent spatial calibration across tools. Furthermore, integrating radiance field-based representations with photogrammetric filtering—combining dense geometric reconstruction with learned radiance-based refinement—could improve both realism and quantitative precision. Ultimately, this study confirms that photogrammetry-based reconstruction, especially using open-source solutions such as COLMAP and Meshroom, represents a robust and cost-effective pathway for quantitative road surface evaluation. The results highlight the growing potential of hybrid photogrammetric–neural frameworks for accurate, sensor-free, and scalable road surface monitoring in future research.